\documentclass{article}

\usepackage{PRIMEarxiv}

\usepackage[utf8]{inputenc} 
\usepackage[T1]{fontenc}    
\usepackage{hyperref}       
\usepackage{url}            
\usepackage{booktabs}       
\usepackage{amsfonts}       
\usepackage{nicefrac}       
\usepackage{microtype}      
\usepackage{lipsum}
\usepackage{fancyhdr}       
\usepackage{graphicx}       
\graphicspath{{media/}}     

\usepackage{subcaption}
\usepackage{algorithm}
\usepackage{algpseudocode}
\usepackage{caption}
\usepackage{subcaption}

\algdef{SE}[DOWHILE]{Do}{doWhile}{\algorithmicdo}[1]{\algorithmicwhile\ #1}%

\title{Towards replicated algorithms}

\author{
  Iztok Fister Jr., Iztok Fister \\
  Faculty of Electrical Engineering and Computer Science \\
  University of Maribor \\
  Slovenia\\
  \texttt{\{iztok.fister1, iztok.fister\}um.si} \\  
}

\begin{document}
\maketitle

\begin{abstract}
The main deficiency of the algorithms running on digital computers nowadays is their inability to change themselves during the execution. In line with this, the paper introduces the so-called replicated algorithms, inspired by the concept of developing a human brain. Similar to the human brain, where the process of thinking is strongly parallel, replicated algorithms, incorporated into a population, are also capable of replicating themselves and solving problems in parallel. They operate as a model for mapping the known input to a known output. In our preliminary study, these algorithms are built as sequences of arithmetic operators, applied for calculating arithmetic expressions, while their behavior showed that they can operate in the condition of open-ended evolution. 
\end{abstract}


\section{Introduction}
\textbf{Phanta Rhei} is probably one of the most important quotes in philosophy that was coined by Heraclitus. The term Phanta Rhei is translated commonly as \textit{everything flows}, i.e. everything is changing, nothing is static. It can easily be translated into different aspects of the universe, e.g. humans are changing through time; different material becomes tired, even if we talk about steel. Everything in the universe is changing.

The concept of Phanta Rhei also opens the possibilities to think about the concept of computer programs. Nowadays, computer programs run on digital computers, and are governed according to von Neuman architecture~\cite{neumann1993first}. The architecture divides the program into an algorithm and data, that are both stored into the primary memory during the program's execution. The algorithm consists of a sequence of program instructions capable of manipulating data, and it is more or less considered as static pieces of source codes that are either compiled to machine code, or interpreted by an interpreter and then executed. The main ingredients of the algorithm do not changing themselves during the execution~\cite{roitblat2020algorithms}. Actually, this lays unchanged into the primary memory and the developer is the only driving force for making changes, fixing bugs, making upgrades, or providing patches. 

Human's behavior is determined by their genetic information encoded into a genotype as a sequence of DNA~\cite{farrell2018computational}. Although the genetic code is unique for all living beings, an expression of genotype (i.e., phenotype) determining the characteristics (i.e., traits) of a human depends on the conditions of the environment~\cite{baeck1996evolutionary}. Moreover, more variants of the genetic programs, based on the same genetic code, solve the same problem in parallel, while the thinking process in the human brain decides what is the most proper for themselves according to its experiences~\cite{eysenck2019cognitive}. Finally, the genetic code is also robust on changes of the environment. This means that this must be resistant to conditions of open-ended evolution, i.e., part of an artificial life, that searches for conditions where evolving systems never settle into a single stable equilibrium~\cite{packard2019open-ended}.

The objective of this paper is to introduce the concept of the so-called replicated algorithms, capable of solving the same problem in parallel. These algorithms are represented as a sequence of building blocks (similar to DNA) enabling mapping of the known input to the known output data. Thus, the building blocks represent clues how to manage input data to produce output, while the sequence encodes an algorithm. Moreover, the input and output data can be modified dynamically during the execution of the algorithms by a user, while they need to be adapted to these modifications automatically. The replicated algorithms combined into a population are capable of being (self-)replicated, according to rules based on some intelligent control similar to the process necessary for developing a human brain~\cite{hiesinger2021self}. 

As a test-bed for verification of the concept, searching for a population of algorithms was selected for solving simple arithmetic expressions that are represented in prefix notation with a known input, and the corresponding result as the output, where the algorithms are presented as a sequence of arithmetic operators for addition, subtraction, multiplication, and integer division. An evolutionary algorithm was developed for an intelligent control that works on the population of algorithms that replicate themselves using predefined variation operators taken from evolutionary computation~\cite{eiben2015introduction}. The preliminary results of the experimental study showed that the proposed replicated algorithms can operate in the conditions of open-ended evolution~\cite{lynch2007evolution} as well.

The main contributions of this paper are as follows:
\begin{itemize}
    \item proposing a new concept of replicated algorithms,
    \item showing the bottlenecks of a paradigm "program = algorithm + data structure",
    \item evaluating the proposed concept in conditions of open-ended evolution.
\end{itemize}

\section{Basic information}
This section reviews background information needed for understanding the topics that follow.

\subsection{Developing a human brain}
The inspiration behind the introduction of the replicated algorithms is found in the operation of the human brain, where a behavior of a particular human is described by a corresponding brain structure. The brain structure prescribes how neurons are connected between each other, while developing the brain is governed by genes, i.e., a sequence of nucleotides in DNA representing the basic units of heredity. However, the genes do not represent any blueprint for structuring the brain, but are carriers of algorithmic information (i.e., intelligent control) for their development.

Indeed, two theories about developing brain exists, as follow~\cite{gottlieb2001individual}:
\begin{itemize}
    \item predetermined: \\
    $\mathrm{Genes}\rightarrow\mathrm{Brain structure}\rightarrow\mathrm{Brain function}\rightarrow\mathrm{Experience}$,
    \item probabilistic: \\
    $\mathrm{Genes}\leftrightarrow\mathrm{Brain structure}\leftrightarrow\mathrm{Brain function}\leftrightarrow\mathrm{Experience}$, 
\end{itemize}
The first theory asserts that genes dictate the structure of the brain, which enables specific functions and thus determines the kind of experience. According to the second theory, the experience can be influenced by the brain structure and expression of genes, and vice versa. This means that the same genetic program in the same environment would not produce an exact copy of behavior. For instance, identical twins have the similar, but not identical brains, and, indeed, react differently in the same situation in the environment~\cite{ward2020student}.

\subsection{Epigenetic}
The genetic code prescribes how to translate information encoded within DNA molecules into proteins. This code is unique for all living beings. Although the translation of the code is fixed, the expression of this is highly dynamic~\cite{ward2020student}. Indeed, the expression of the genetic code is influenced by the environment. The phenomenon is also termed as epigenetic~\cite{bird2007perceptions}.

In general, the static genetic program may have a different expression. In the sense of computer programs, there are many algorithm's variants for how to transform the input data into the output. Usually, in computer theory, these are distinguished according to their efficiency in the sense of time and space~\cite{garey1979computers}.

\subsection{Artificial life}
Artificial life refers to a scientific field concerned with the strategy of artificial systems that mimics or manifests the property of living systems during the long-term evolution~\cite{adami1998introduction}. Interest in this domain has increased with moving the nature-inspired algorithms into the hardware~\cite{eiben2015from}. This process was raised by the emergence of domains, like Evolutionary Robotics (ER) and Swarm Robotics (SR). Both domains deal with the problems of how to avoid a loss of population diversity that leads the evolutionary search process to get stuck into the local optima, and, consequently, terminate this prematurely~\cite{doncieux2019novelty}. 

\section{Concept of replicated algorithms}
A modeling problem, solved in the preliminary study, can be defined formally as follows: Let a set of building blocks for creating a universal algorithm and a mapping input data into output $I\mapsto O$ be defined. Then, the task of the modeling is to find a population of such algorithms that would be able to solve a given problem (i.e., $I\mapsto O$) in each moment, could be adapted to new conditions of an environment, and, thus, resist the conditions of open-ended evolution.  An inspiration for the design of such a population was found in challenges of the Origin-Of-Life (OOL), asserting that living cells can self-replicate, which, in turn, self-replicate the self-replicator, ad infinitum~\cite{lo2020evolution}. Thereby, the concept of self-replication is widened to the concept of replication, where the computer algorithm can replicate itself, the copies of which could replicate themselves, and so on. In the sense of EC, the concept of self-replication is closer to asexual mutation as an origin of the new algorithm, while the concept replication is broader, and refers to the bi-sexual crossover as well as asexual mutation.

The design of the replicated algorithms is based on an intelligent control that controls a population of algorithms capable of replicating themselves by using the variation operators (i.e., crossover and mutation), and thus enables them to operate in conditions of open-ended evolution. In summary, the intelligent control is implemented as an evolutionary algorithm working on population of individuals represented as variable sized sequences of steps, in which the given problem can be solved.

Indeed, the characteristics of the replicated algorithms are as follows:
\begin{itemize}
    \item each individual is represented as a variable sequence of building blocks similar to the genetic code,
    \item the knowledge is inherited from the sequences of building blocks,
    \item program is an independent unit, standalone, able to be ported to different machines,
    \item they are able to operate in conditions of open-ended evolution.
\end{itemize}

\begin{figure}[htb]
    \centering
    \includegraphics[scale=.54]{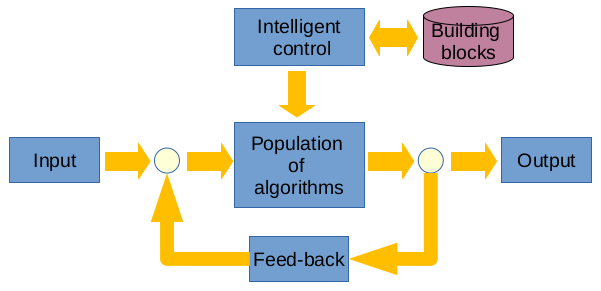}
    \caption{Concept of replicated algorithms.}
    \label{fig:my_label}
\end{figure}

The concept of the replicated algorithms is illustrated in Fig.~\ref{fig:my_label},
from which it can be seen that the intelligent control controls a population of algorithms presented as a variable sized sequence of building blocks (i.e., clues) transforming the known input data to the known output. Thus, the population of algorithms acts as a feed-back system, where the fitness function of each algorithm is presented as an error value measuring the difference between the desired output and effective result of this. 

\section{Experiments and results}
The goal of the experimental work was to show that the replicated algorithms are capable of solving real-world problems. Furthermore, these algorithms can be replicated during the execution and thus evolve from the simple organization far away from the given goal (i.e., the proper result) toward a complex organization capable of achieving this goal. Using intelligent control, the population of algorithms is capable to resist in conditions of open-ended evolution as well.

In our preliminary study, the task of the intelligent control was to build an algorithm able to calculate the exact value of an arithmetical expression from the operands entered as an input. Obviously, the algorithm consists of a sequence of simple operators, like addition, subtraction, multiplication, and division that are represented in postfix notation, where operator appears after the operands. The main advantage of this representation is implicit to considering the parentheses, which leads to a faster calculation. 

The Evolutionary Control Algorithm (ECA) played the role of the intelligent control. Two population models were employed by the ECA: a generational, and a steady-state~\cite{eiben2015introduction}. The former exchanges the whole population with the best individuals between the original and the mating pool populations, while the latter exchanges only the part of the original population, i.e., the generational gap. In the last model, the algorithm can also be changed in the case that the new formed individual from the mating pool is worse than the original one according to their fitness value (i.e., the non-elitist selection). Consequently, the non-elitist selection prevents the evolution process from getting stuck into the local optima, and thus creates conditions for open-ended evolution.

The other parameters of the ECA were set during the experiments as illustrated in Table~\ref{lab:1}, where $D_{init}$ limits the maximum algorithm's size in the initial population, normally, determined randomly in the interval $[1,D_{init}]$, $Np$ presents the population size, $nGen$ the maximum number of generations, $p_{mut}$ the probability of mutation, and the generation gap $1/10$, denoting that one of $Np=10$ individuals in the steady-state model will be changed using the crossover operator in each generation. However, the crossover probability $p_c$ is set to one, if the generational population model is selected. In both cases, the one-point crossover is employed on variable-sized individuals.
\begin{table}[htb]
\caption{Evolutionary control algorithm setting.}
\label{lab:1}
\begin{center}
\begin{tabular}{ l l r }
Parameter & Symbol & Value \\
\hline
 Maximum initial algorithm's size  & $D_{init}$ & 10 \\ 
 Population size & $Np$ & 10 \\  
 Generation number & $nGen$ & 1,000,000 \\
 Mutation probability & $p_m$ & 0.1 \\  
 Generational gap (Steady-state) & $\lambda/Np$ & 1/10 \\
 Crossover probability (Generational) & $p_c$ & 1.0 \\  
  \hline
\end{tabular}
\end{center}
\end{table}
In a generational population model, the elitist selection is employed, where the best between the trial and target individual survives into the next generation. Let us mention that the set of input operands was defined as $I=\{10,20,30\}$ and the result as $O=\{100\}$.

The results of the ECA using the generational population model and the elitist selection are illustrated in Table~\ref{lab:1},
\begin{table}[htb]
\caption{The results of the ECA using generational population model and elitist selection.}
\label{lab:1}
\begin{center}
\begin{tabular}{ l l r }
Nr. & Program & Fitness \\
\hline
1 & $* 10~10$ & 0 \\
2 & $* 10~10$ & 0 \\
3 & $* 10~10$ & 0 \\
4 & $* 10~10$ & 0 \\
5 & $* 10~10$ & 0 \\
6 & $* 10~10$ & 0 \\
7 & $* 10~10$ & 0 \\
8 & $* 10~10$ & 0 \\
9 & $* 10~10$ & 0 \\
10 & $* 10~10$ & 0 \\
 \hline
\end{tabular}
\end{center}
\end{table}
from which it can be seen that the population of algorithms converged fast to the optimal expression written in the infix notation as follows:
\begin{equation}
    10*10=100.
    \label{eq:1}
\end{equation}
In contrast, the results of the ECA using the steady-state population model and non-elitist selection, as depicted in Table~\ref{lab:2}, 
\begin{table}[htb]
\caption{The results of the ECA using the steady-state population model and non-elitist selection.}
\label{lab:2}
\begin{center}
\begin{tabular}{ l l r }
Nr. & Program & Fitness \\
\hline
1 & $+ / 20~30 / * - 30~20~10 / * 20~30~10$ & 99 \\
2 & $ + / * 30~30~10 - 30~20$  & 0 \\
3 & $+ / * 30~20~10 / * 20~20~10$ & 0 \\
4 & $+ / 20~30 / * * 10~20~10~20$ & 0 \\
5 & $+ / / * 30~30~10 / 10~10~10$ & 0 \\
6 & $+ / / * 30~30~10 / + 20~30 / * 30~10~10 - 30~20$ & 0 \\
7 & $+ / * 30~30~10 - 30~20~30~10~10$ & 0 \\
8 & $+ / / 30~10~10 / * * 10~20~10~20$ & 0 \\
9 & $+ / / * 30~30~10 / + 20~30~30~10$ & 0 \\
10 & $+ / / 30~10~10 / * * 10~10~30~30$ & 0 \\
 \hline
\end{tabular}
\end{center}
\end{table}
did not converge to the same optimal algorithm, but the ECA maintained the population of different algorithms obtaining, almost in each independent run, the proper solution. This population shows that the ECA using the steady state model and non-elitist selection can also persist in conditions of open-ended evolution.

\begin{figure}[htb]
\begin{subfigure}[c]{0.2\textwidth}
    \centering
    \includegraphics[scale=.48]{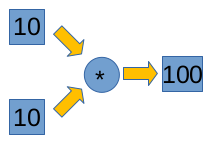}
    \caption{Term 1.}
\end{subfigure}
\begin{subfigure}[c]{0.25\textwidth}
    \centering      
    \includegraphics[scale=.48]{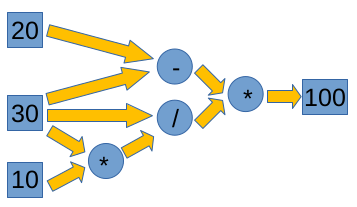}
    \caption{Term 2.}
\end{subfigure}
\caption{Two samples of terms from a population of programs.}
\label{fig:1}
\end{figure}

Two examples of terms in population of programs maintained by the ECA are illustrated in Fig.~\ref{fig:1},
where the figure is divided into two diagrams, and both diagrams consist of input operands, the algorithm for transforming the input operands to the output result represented as numbers within squares, and the algorithm represented as a sequence of circles with corresponding arithmetic operators. The diagram in Fig.~\ref{fig:1}(a) depicts the optimal program (Eq.~\ref{eq:1}), while the diagram in Fig.~\ref{fig:1}(b) shows an example of the arithmetic expression that produces the proper result in open-ended evolution.

\subsection{Discussion}
The results of the experiments showed that the population of replicated algorithms can solve problems in parallel. Interestingly, the population of replicated algorithms under ECA using the generational population model converged to the optimal algorithm (Eq.~\ref{eq:1}) in the sense of calculation without any external selection pressure. On the other hand, the same population under ECA using the steady-state population model maintained high population diversity in most independent runs. Thereby, two behaviors can be observed, i.e., the population either converges toward the optimal solution, or starts to oscillate between the population members, and thus created conditions for open-ended evolution.

As a result, the necessary conditions for open-ended evolution, as observed in our study, can be reviewed as follows:
\begin{itemize}
    \item the small population-size,
    \item the variable-sized representation of individuals,
    \item the steady-state population model.
\end{itemize}

The von Neumann architecture supposes that the algorithm loaded into memory stays unchanged during the execution, and it can be changed only when the current instance of the program terminates, and the new version is made and loaded into the primary memory again. The replicated algorithms, solving the problems in parallel, could be adapted to the new condition of environment automatically. 

However, two drawbacks may be connected with the current version: (1) the fixed population size, and (2) external intelligent control introduced by the ECA. Fortunately, both issues could be solved by changing the population members into intelligent agents, while the intelligent control might be transferred from ECA to the intelligent multi-agent system. 

\section{Conclusion}
This study introduces the concept of replicated algorithms capable of solving modeling problems in parallel. They work under the control of an intelligent control (in our case ECA) that transforms the known input data into the proper results. Thus, the algorithms in the population are presented as a sequence of orders, and each of them usually represents a different solution to the problem. The results of the experiments by constructing the algorithm for controlling arithmetic expressions showed that some configurations of them can also be employed in conditions of open-ended evolution.

Primarily, the potential directions for the future work are obviously eliminating the exposed drawbacks. This means that, at first, the central intelligent control should be distributed between the intelligent agents capable of making decisions based on local information only. Then, the multi-agent system should be implemented with a variable number of agents. Secondly, conditions for open-ended evolution might also be discovered in detail.

\bibliographystyle{unsrt}  
\bibliography{references}

\end{document}